\documentclass[letterpaper]{article} 
\usepackage{aaai24}  
\usepackage{times}  
\usepackage{helvet}  
\usepackage{courier}  
\usepackage[hyphens]{url}  
\usepackage{graphicx} 
\usepackage{natbib}  
\usepackage{caption} 
\usepackage{algorithm}
\usepackage{algorithmic}

\usepackage{newfloat}
\usepackage{listings}

\usepackage{epsfig}
\usepackage{amsmath}
\usepackage{amssymb}
\usepackage[caption=false]{subfig}

\usepackage{tabularx}
\usepackage{enumitem}
\usepackage{booktabs, multirow} 
\usepackage{soul} 
\usepackage[table,dvipsnames]{xcolor} 

\usepackage[capitalize]{cleveref}
\usepackage{ragged2e}
\usepackage{blindtext}
\usepackage{xspace}
\usepackage{dsfont}
\usepackage{enumitem}

\DeclareCaptionStyle{ruled}{labelfont=normalfont,labelsep=colon,strut=off} 
\floatstyle{ruled}
\newfloat{listing}{tb}{lst}{}
\floatname{listing}{Listing}
\title{Learning Pseudo-Labeler beyond Noun Concepts \\ for Open-Vocabulary Object Detection}
\author{
    Sunghun Kang\textsuperscript{\rm 1,2}, Junbum Cha\textsuperscript{\rm 1}, Jonghwan Mun\textsuperscript{\rm 1}, Byungseok Roh\textsuperscript{\rm 1}, Chang D. Yoo\textsuperscript{\rm 2}
}
\affiliations{
    \textsuperscript{\rm 1} Kakao Brain\\
    \textsuperscript{\rm 2} Korea Advanced Institute of Science and Technology\\
    \{sunghun.kang, junbum.cha, jason.mun, peter.roh\}@kakaobrain.com \\
    cd\_yoo@kaist.ac.kr
}

\usepackage[capitalize]{cleveref}
\crefname{figure}{Figure}{Figures}
\crefname{section}{Section}{Sections}
\crefname{table}{Table}{Tables}

\newcommand{\eg}{\emph{e.g}.}
\newcommand{\ie}{\emph{i.e}.}
\newcommand{\etal}{\emph{et al}.}

\newcommand{\fullmethod}{\textbf{P}seudo-\textbf{L}abeling for \textbf{A}rbitrary \textbf{C}oncepts\xspace}
\newcommand{\method}{PLAC\xspace}

\newcommand{\bridging}{Rasheed \etal{}}

\newcommand{\apr}{AP$_\text{r}$}

\newcolumntype{x}[1]{>{\centering\arraybackslash}p{#1pt}}
\newcommand{\tablestyle}[2]{\setlength{\tabcolsep}{#1}\renewcommand{\arraystretch}{#2}\centering\small}

\definecolor{light-gray}{gray}{0.50}

\begin{document}
\maketitle

\begin{abstract}
Open-vocabulary object detection (OVOD) has recently gained significant attention as a crucial step toward achieving human-like visual intelligence. Existing OVOD methods extend target vocabulary from pre-defined categories to open-world by transferring knowledge of arbitrary concepts from vision-language pre-training models to the detectors. While previous methods have shown remarkable successes, they suffer from indirect supervision or limited transferable concepts. In this paper, we propose a simple yet effective method to directly learn region-text alignment for arbitrary concepts. Specifically, the proposed method aims to learn arbitrary image-to-text mapping for pseudo-labeling of arbitrary concepts, named \fullmethod (\method). 
The proposed method shows competitive performance on the standard OVOD benchmark for noun concepts and a large improvement on referring expression comprehension benchmark for arbitrary concepts.
\end{abstract}

\section{Introduction}
\begin{figure}[t]
    \centering
    \subfloat[Previous pseudo-labeling]{
        \includegraphics[width=0.47\linewidth]{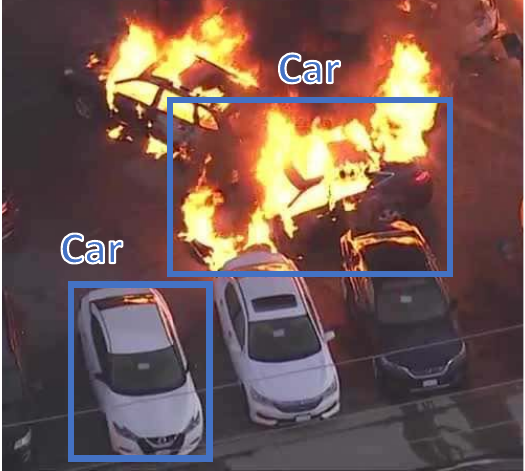}
        \label{fig:highlight_plac_a}
    }
    \subfloat[PLAC (Ours)]{
        \includegraphics[width=0.47\linewidth]{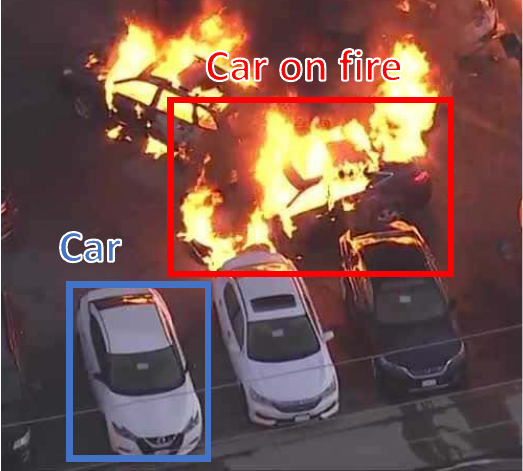}
        \label{fig:highlight_plac_b}
    }
    \vspace{-0.25cm}
    \caption{
        Comparison between existing noun-based pseudo-labeling and the proposed \method. (a) noun-based methods suffer from pseudo-labeling complex concepts, \eg, ``car on fire'', (b) the proposed method is able to generate pseudo-labels for arbitrary concepts beyond noun concepts.
    }
    \label{fig:highlight_plac}
\end{figure}

\begin{figure*}[ht!]
    \begin{center}
    \includegraphics[width=1.0\linewidth]{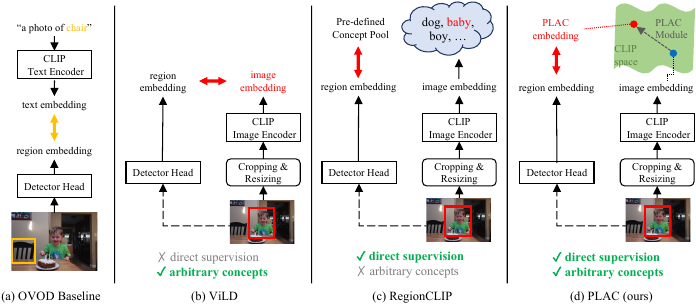}
    \end{center}
    \vspace{-0.35cm}
    \caption{
    Conceptual differences of supervision in existing open-vocabulary detection methods. (a) OVOD baseline is trained using pairs of (region, base category) as supervision. (b) ViLD uses CLIP image embeddings, which is indirect supervision for OVOD. (c) RegionCLIP leverages the concept pool that is a set of \emph{pre-defined nouns}. (d) The proposed method utilizes the \method that translates image embeddings to the pseudo embeddings in CLIP embedding space.}
    \label{fig:concept_comparison}
    \vspace{-0.35cm}
\end{figure*}

The recent breakthrough in open-world visual recognition \cite{radford2021clip} prompts the object detection community to shift its focus to a new area, called open-vocabulary object detection (OVOD)---aiming at learning to detect arbitrary visual concepts beyond the pre-defined training classes  \cite{gu2022vild,zhong2022regionclip,Hanoona2022Bridging,lin2022vldet,minderer2022owlvit,zareian2021ovrcnn,zhou2022detic}.
The capability of OVOD to detect arbitrary concepts beyond pre-defined categories unlocks various novel possibilities for real-world applications. For example, detecting ``car on fire'' (\cref{fig:highlight_plac}) typically requires extensive data gathering and model training in traditional object detection scenarios. However, OVOD aims to detect arbitrary concepts including ``car on fire'' without any additional data gathering or model training.
A major approach for OVOD is transferring knowledge of arbitrary concepts from open-world classification models (\eg, 
CLIP~\cite{radford2021clip}) to detection models. Specifically, it extends the target vocabulary from close-set to open-set by leveraging textual descriptions of classes. As illustrated in \cref{fig:concept_comparison}a, an open-vocabulary detector predicts the category of a given proposal by region-text alignment, \ie, alignment between the estimated region embeddings and text embeddings from the target class descriptions. 

The main challenge in OVOD is how we can effectively transfer the knowledge of an open-vocabulary classifier (\ie, CLIP) to a detector. 
For example, ViLD \cite{gu2022vild} attempts to address this challenge by employing knowledge distillation between detector region embeddings and CLIP image embeddings, as illustrated in \cref{fig:concept_comparison}b.
However, such distillation may not be optimal for OVOD.
This is because OVOD models require region-text alignment at test time, but, the distillation only optimizes alignment between region embeddings and CLIP image embeddings, not CLIP text embeddings.
Such indirect optimization brings discrepancy between the training and testing phases, limiting the OVOD performance.

RegionCLIP \cite{zhong2022regionclip} addresses this problem by introducing pseudo-labels instead of indirect supervision through image embedding distillation, allowing for direct region-text alignment learning (\cref{fig:concept_comparison}c).
To obtain pseudo-labels, RegionCLIP builds a concept pool using frequently occurring nouns in image captions.
It extracts pseudo-labels by performing classification using the concept pool and CLIP.
Although this approach has shown remarkable success in OVOD by enabling direct optimization of region-text alignment (\ie, region-noun alignment), it still has limitations in transferring knowledge of CLIP beyond single-noun concepts.
Thus, we believe that a new pseudo-labeling method---beyond noun concepts---is required to truly detect arbitrary concepts, \eg, ``car on fire'' in \cref{fig:highlight_plac}.

\vspace{-0.06cm}
In this paper, we propose a simple yet effective method, named \fullmethod{} (\method), 
that leverages pseudo-labeling to directly optimize region-text alignment while also expanding the transferable knowledge of CLIP beyond noun concepts.
Instead of constructing a noun concept pool, we train a pseudo-labeler using image-text pairs \cite{sharma2018cc3m}, which learns arbitrary image-to-text mapping on CLIP embedding space. Note that we train the pseudo-labeler in the latent space since our target label is text embedding of CLIP. That is, we aim to learn $e^t=\theta(e^i)$ where $e^t$ and $e^i$ represent CLIP text embedding and CLIP image embedding, respectively, and $\theta$ denotes the pseudo-labeler. 
To generate pseudo-labels, we first extract image embeddings by using a pre-trained CLIP image encoder on pre-extracted region proposals. Then, we apply a learned pseudo-labeler to extract pseudo-labels, \ie, pseudo text embeddings.
The overview of this process is illustrated in \cref{fig:concept_comparison}d.
\method{} enables the OVOD model to learn arbitrary concepts beyond noun concepts through direct supervision for region-text alignment.
In addition, we also propose a two-stage bipartite matching strategy to effectively address uncertainty of pseudo-labels.
We employ two different benchmarks to evaluate the capabilities of detecting noun and beyond-noun concepts: LVIS and referring expression comprehension (REC), respectively. 
The proposed method significantly improves performance on both benchmarks compared to the na\"ive baseline (\cref{fig:concept_comparison}a) with base-only training.
Notably, the superior performance in REC highlights the effectiveness of our method in enhancing the capability of detecting beyond noun concepts.

Our main contributions are summarized as follows:
\begin{itemize}[leftmargin=.7cm,noitemsep,nosep]
    \item
        We propose a simple yet effective pseudo-labeling method for OVOD, named \method, which mainly aims to transfer the knowledge of CLIP for arbitrary concepts beyond noun concepts.
    \item
        We propose the two-stage matching strategy to effectively leverage the full advantage of both annotated labels and pseudo-labels.
    \item 
        \method demonstrates greater capability of detecting arbitrary concepts in the REC benchmark, while its performance is comparable to existing OVOD methods in the standard noun-based benchmark.
\end{itemize}

\section{Related Works}
\subsection{Vision-Language Pre-training}
Recent advances in the vision-language pre-training (VLP) field usher in extending the vision task beyond traditional close-set prediction on the pre-defined categories to the open-world \cite{radford2021clip,gu2022vild,zhong2022regionclip,xu2022groupvit,cha2022tcl}. 
These VLP models are capable of handling arbitrary concepts by learning vast knowledge from near-infinite web-crawled image-text pairs \cite{sharma2018cc3m,kakaobrain2022coyo-700m,schuhmann2021laion}.
Notably, the pioneering work of Contrastive Language-Image Pre-training (CLIP) demonstrates the zero-shot image classification capability in the open-world setting by learning image-text alignment.
The advent of open-world image classification models enables open-world settings in various fields such as object detection \cite{gu2022vild, zhong2022regionclip} or semantic segmentation \cite{cha2022tcl,xu2022groupvit}.

\subsection{Open-Vocabulary Object Detection (OVOD)}
The success of VLP leads the object detection field to the open-world scenario, \ie, OVOD. 
The goal of OVOD is to localize arbitrary concepts beyond pre-defined categories, thereby opening up novel possibilities for real-world applications.
The primary approach for OVOD involves leveraging existing object detection datasets to learn localization capabilities and extending target concepts through knowledge transfer from VLP models.
\cref{fig:concept_comparison}a illustrates how open-vocabulary detectors expand the target vocabulary using text encoders of pre-trained VLP models. Specifically, these detectors replace the close-set classifier with the nearest neighbor classifier, utilizing text embeddings from the textual description of target classes. As large-scale VLP models can encode arbitrary textual concepts, the detector can address arbitrary concepts only if it can encode arbitrary visual concepts into region embeddings.

Existing OVOD methods tackle this problem---how to encode arbitrary visual concepts---by transferring knowledge from the image encoder of pre-trained VLP models to the detector in various ways \cite{gu2022vild,zhong2022regionclip,Hanoona2022Bridging,lin2022vldet,minderer2022owlvit,zareian2021ovrcnn,zang2022ovdetr,zhou2022detic,li2022glip,gao2022open,zhao2022exploiting}. 
One approach is to simply employ the image encoder of pre-trained VLP models as a backbone of the detector \cite{minderer2022owlvit,kuo2022fvlm}. While simple, this approach limits the backbone types and tends to require a large base vocabulary.
Another approach is employing a knowledge distillation approach to transfer knowledge of pre-trained VLP models \cite{gu2022vild,Hanoona2022Bridging}.
ViLD \cite{gu2022vild} proposes embedding distillation between the detector region embeddings and CLIP image embeddings. \bridging{} \cite{Hanoona2022Bridging} improve the distillation by considering the gap between object and image-level embeddings.
However, these methods suffer from train-test discrepancy due to the lack of direct supervision on region-text alignment.
Pseudo-labeling methods are not affected by this discrepancy since they directly learn region-text alignment \cite{zhong2022regionclip,lin2022vldet}. 
Although this approach has shown remarkable success, it relies on noun concepts, presenting another challenge.
In particular, they extract frequently occurring nouns from image captions using a rule-based off-the-shelf parser \cite{schuster2015scenegraphparser}, and use them as candidates for pseudo-labels.
Thus, the transferable knowledge is bounded to the performance of the off-the-shelf parser and restricted to noun concepts.
In this paper, we propose a novel pseudo-labeling method, \method, which enables to transfer knowledge of arbitrary concepts without being restricted to a rule-based parser or noun concepts.

\section{Proposed Methods}
\begin{figure*}[!t]
    \begin{center}
        \includegraphics[width=1.0\linewidth]{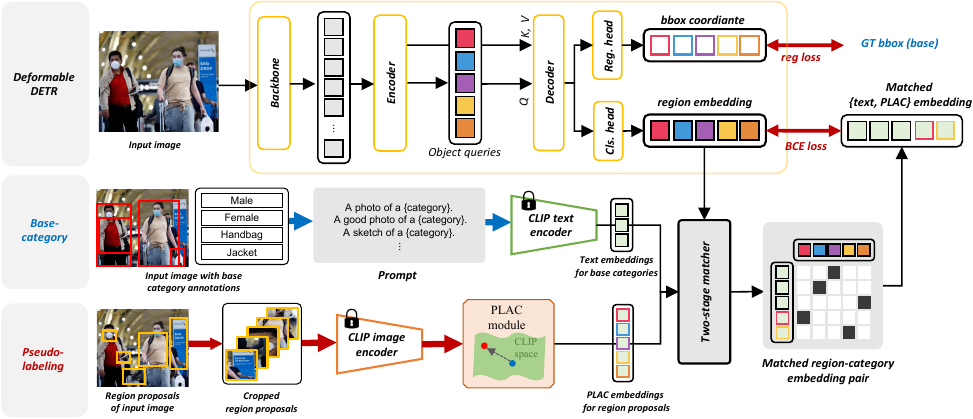}
    \end{center}
    \vspace{-0.3cm}
    \caption{
    An overview of the proposed \method-based OVOD. 
    The base categories and pre-extracted region proposals are encoded using the frozen CLIP text encoder and image encoder, respectively.
    The CLIP image embeddings of the given proposal regions are then translated into \method embeddings using the proposed \method module.
    The two-stage matcher matches the region embeddings predicted by Deformable DETR with both the text embeddings for base categories and \method embeddings for region proposals.
    In the matching process, we introduce a base-first matching strategy; it matches region proposals with base text embeddings first to minimize the uncertainty of the pseudo-labels, before proceeding to match with \method embeddings.
    }
    \label{fig:overall_method}
    \vspace{-0.3cm}
\end{figure*}
Object detection is a task of localizing and classifying all objects in the images $x^i$.
Formally, for the pre-defined (base) categories $\mathbb{C}_{\text{base}}$, the conventional object detector is trained to extract region embeddings $o_i$, and then identify bounding box coordinates $b_i$ and associated object names $c_i \in \mathbb{C}_{\text{base}}$ from $o_i$.
Open-Vocabulary Object Detection (OVOD) aims to extend the target vocabulary from close-set (\ie, base categories $\mathbb{C}_{\text{base}}$ in training data) to open-set (\ie, novel categories $\mathbb{C}_{\text{novel}}$ during testing).

OVOD is mainly achieved by transferring the knowledge of a pre-trained open-world classification model, \ie, CLIP~\cite{radford2021clip}, which has knowledge of a large vocabulary ($\mathbb{C}_{\text{open}}$\footnote{It is expected that the large vocabulary $\mathbb{C}_{\text{open}}$ partially or fully contains visual concepts in $\mathbb{C}_{\text{base}}$ or $\mathbb{C}_{\text{novel}}$.}) from image-text paired data.
Generally, as shown in \cref{fig:concept_comparison}a, leveraging CLIP text embeddings of target object labels as supervision enables OVOD.
Then, pseudo-labeling of noun concepts~\cite{zhong2022regionclip} using CLIP is further introduced to provide knowledge beyond $\mathbb{C}_{\text{base}}$.
However, we observe that such knowledge transfer of only nouns is not optimal for OVOD.

For the task, as shown in \cref{fig:overall_method}, we propose a novel OVOD framework with the \fullmethod (\method) module.
The \method module is in charge of generating pseudo-labels of arbitrary concepts ($\mathbb{C}_{\text{open}}$) for region candidates (or proposals).
Then, in addition to detection annotations for $\mathbb{C}_{\text{base}}$, the extracted pseudo-labels for $\mathbb{C}_{\text{open}}$ are used to train a detector, \ie, Deformable DETR~\cite{zhu2021deformabledetr}.
Note that we further introduce a two-stage matching based learning scheme to facilitate use of the extracted pseudo-labels during training detector.

\subsection{\fullmethod}
\begin{figure}[!t]
    \centering
    \includegraphics[width=1.0\linewidth]{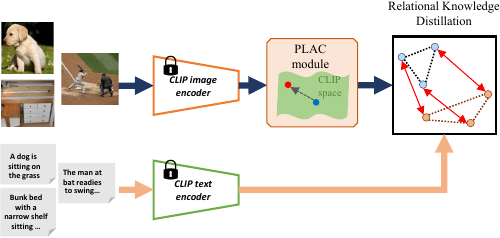}
    \vspace{-0.45cm}
    \caption{
    Learning \method module. 
    Using image-text paired data, \method module is trained to translate CLIP image embeddings into its corresponding CLIP text embeddings.
    RKD is also employed to learn the structural relations between a set of image-text pairs beyond point-wise mapping.
    }
    \vspace{-0.2cm}
    \label{fig:PLAC}
\end{figure}

Our key component is how we can extract beneficial pseudo-labels, which properly contain knowledge of CLIP over arbitrary concepts, from regions of images.
This raises a penetrating question: \textit{what is the knowledge of CLIP for a given image?}
We believe it is a CLIP text embedding that has the highest similarity with a CLIP image embedding of the given image.
With this motivation, we propose \fullmethod (\method) that translates images into their CLIP text embeddings.
More specifically, \method learns a mapping function of CLIP image embedding $\rightarrow$ CLIP text embedding, named \method embedding (\cref{fig:PLAC}).
The major difference compared to concept-pool based pseudo-labeling approaches~\cite{zhong2022regionclip, lin2022vldet}, which assign the given region embedding to one of the pre-defined noun concepts, is that the proposed \method predicts a vector lie on the CLIP text embedding space. The \method does not require pre-defined concepts and can represent any arbitrary concepts beyond nouns learned in CLIP.
Note that we present an interpretation of PLAC embeddings in the Appendix, showing PLAC embeddings capture diverse semantics of images beyond simple nouns.

The mapping function from CLIP image embedding to CLIP text embedding, called PLAC module, is defined by a three-layer multi-layer perceptron with \texttt{gelu}~\cite{hendrycks2016gelu} activation.
Note that we use image-text paired data to train the PLAC module, just like when CLIP was trained.
Given a image-text pair, $(x^i, x^t)$, let us denote $e^i = \text{CLIP}_{\text{I}} (x^i)$ and $e^t = \text{CLIP}_{\text{T}} (x^t)$ by CLIP image and text embeddings, respectively.
PLAC module produces PLAC embedding $e^{p}$ from CLIP image embedding:
\begin{equation}
    e^{p} = \text{PLAC}(e^i) = \text{PLAC}(\text{CLIP}_{\text{I}}(x^i)).
\end{equation}
Then, the PLAC module is trained to mimic CLIP text embedding with two losses. 
First, we use Mean-Squared-Error (MSE) loss between PLAC and CLIP text embeddings, 
\begin{equation}
    \mathcal{L}_{\text{MSE}} = \frac{1}{|\mathcal{X}|} \sum_{(x^i_k, x^t_k) \in \mathcal{X} } ||e^t_k - e^p_k||_2^2,
\end{equation}
where $\mathcal{X}$ means a mini-batch of image-text pairs.

Second, we adopt Relational Knowledge Distillation loss (RKD)~\cite{park2019rkd} to facilitate the transfer of mutual relationships between CLIP text embeddings ($e^t$) and \method embeddings ($e^p$) by preserving the same pairwise L2 distances within each set of embeddings as follows:
\begin{equation}
    \mathcal{L}_\text{RKD} = \sum_{ \{ (x^i_j, x^i_k), (x^t_j, x^t_k) \} \in \mathcal{X}^2} 
        l_\delta \left( \psi (e^t_j, e^t_k), \psi (e^p_j, e^p_k)) \right),
\end{equation}
where $l_\delta$ is a  function to compute $\texttt{smooth-L1}$ loss between given two values. $\psi(e_j, e_k) = \frac{1}{\mu} || e_j - e_k||_2$ is a normalized L2 distance function where a normalization factor $\mu$ is derived by $\mu = \frac{1}{|\mathcal{X}^2|} \sum_{(e_j, e_k) \in \mathbf{e}^2} ||e_j - e_k||_2$.
$\mathcal{X}^2$ and $\mathbf{e}^2$ are sets of tuples of distinct data examples and embeddings, respectively, \ie, $\mathcal{X}^2 =  \{(x_i, x_j) | i \neq j \}$, $\mathbf{e}^2 = \{e_i, e_j | i \neq j \}$.

The overall objective for PLAC module is defined by
\begin{eqnarray}
    \mathcal{L}_{\text{\method}} = \mathcal{L}_{\text{MSE}} + \lambda_{\text{RKD}} \cdot \mathcal{L}_{\text{RKD}},
\end{eqnarray}
where $\lambda_\text{RKD}$ is a hyperparameter for controlling the effectiveness of RKD.

\paragraph{Extraction of pseudo-labels for OVOD.}
We use the off-the-shelf Region Proposal Network (RPN) to extract region proposals from the detection dataset where RPN is trained using only base category annotations.
Note that, to prevent competition with the base objects annotation, we discard region proposals that overlap significantly with annotated objects or have low objectness scores.
The PLAC embeddings $e^p$ are extracted from remaining region proposal boxes $b_\text{RPN}$:
\begin{eqnarray}
    e^p = \text{PLAC} (\text{CLIP}_\text{I} (\text{Crop} (x^i, b_\text{RPN}))),
\end{eqnarray}
where $\text{Crop}(x^i, b_\text{RPN})$ represents the cropped image with $b_\text{RPN}$ from input image $x^i$.
Then, the extracted PLAC embeddings for individual region proposals are used as pseudo-labels when training the OVOD model.

\begin{figure}
\begin{center}
\includegraphics[width=1.0\linewidth]{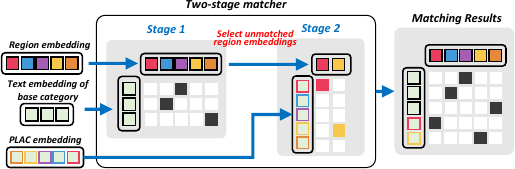}
\end{center}
   \caption{Two-stage matching. At stage 1, the bipartite matching between object queries and text embedding of base categories proceeded. At stage 2, the unmatched object queries are matched with the \method embeddings. Each matching is conducted using Hungarian algorithm.}
\label{fig:matcher}
\end{figure}

\subsection{Deformable DETR with Two-Stage Matching}
\paragraph{Deformable DETR.}
We first briefly recap Deformable DETR~\cite{zhu2021deformabledetr}. 
Deformable DETR is composed of a backbone and a transformer encoder-decoder. The backbone is responsible for extracting the spatial embeddings of a given image. 
The transformer encoder enhances spatial embeddings to gather contextual information via self-attention.
The transformer decoder takes learnable object queries, refines them using spatial features (by the encoder) via cross-attention, and then consequently generates region embedding $o \in \mathcal{O}$ for individual object queries.
The classifier and box regressor heads predict the category and their bounding box, respectively, for each regional embedding $o$.
During training, the Hungarian algorithm is utilized to perform a bipartite matching between the set of predictions and their corresponding ground-truth labels. 

\paragraph{Deformable DETR for OVOD.}
To generalize the Deformable DETR for open-vocabulary object detection, one intuitive way is reformulating the classification head to category agnostic classifier using CLIP.
For $i$-th region embedding $o_i$, the probability $p_{i,c}$ for the arbitrary concept $c \in \mathbb{C}_\text{open}$ is given by:
\begin{equation}
    p_{i,c} = \sigma \left( \alpha \cdot (e^t_c)^\top o_i + \beta \right),
    \label{eq:ovod_prob}
\end{equation}
where $\sigma$ and $e^t_c$ indicate Sigmoid function and CLIP text embedding for arbitrary concept $c$.
$\alpha$ and $\beta$ are adjusting factors to scale and shift logit values, respectively.
Note that all experiments are conducted with $\alpha = 25$ and $\beta = -0.25$.

\paragraph{Two-stage matching.}
Given an image, there are two different types of labels: (1) a set of text embeddings for the annotated base categories, and (2) pseudo-labels of \method embeddings for region proposals $b_{\text{RPN}}$ given by region proposal network.
One straightforward way assigning labels to region embeddings is considering two types of labels with the same importance;
that is, a single-stage matching can be conducted between region embeddings and a union of two types of labels.

However, as presented in Table~\ref{table:matching_strategy}, we observe that such single-stage matching hinders the learning for OVOD;
we conjecture that uncertain and noisy psueo-labels sometimes hinder learning of human-annotated base labels.
Inspired by this issue, we introduce two-stage matching, which is illustrated in \cref{fig:matcher}.
In the first stage, region embeddings are first matched with a set of CLIP text embeddings for the annotated base categories to guarantee the learning of base classes.
In the second stage, we perform matching between pseudo-labels (\ie, PLAC embeddings) and unmatched remaining region embeddings.
OVOD model is trained to increase the similarity (\ie, $p$ in \cref{eq:ovod_prob}) for the matched pairs.
Note that the box regression loss is only applied to the region embeddings matched with base categories.

\begin{table*}[t]
    \centering
    \small
    \scalebox{1.0}{
        \begin{tabular}{l|llc|l|cccc}
        \toprule
        Method                    & Detector     & Backbone & \#Params & Annotation & AP$_\text{r}$ & AP$_\text{c}$ & AP$_\text{f}$ & AP    \\ 
        \midrule \midrule
        \multicolumn{9}{l}{\textit{Controlled comparison under backbone size $< 30$M parameters}} \\
        \midrule
        ViLD \cite{gu2022vild}                          & Mask RCNN    & RN50 & 22M   & Box+Mask    & 16.7 & 26.5 & 34.2 & 27.8  \\
        RegionCLIP \cite{zhong2022regionclip}           & Faster RCNN  & RN50 & 22M   & Box+Mask    & 17.1 & 27.4 & 34.0 & 28.2  \\
        Rasheed \etal{}$^*$ \cite{Hanoona2022Bridging}  & Mask RCNN    & RN50 & 22M   & Box+Mask    & 21.6 & 26.3 & 31.1 & 27.4  \\
        VLDet$^*$ \cite{lin2022vldet}                   & CenterNet V2 & RN50 & 22M   & Box+Mask     & 22.9 & 32.8 & 38.7 & 33.4  \\
        \color{gray} Base-only$^\dagger$ \cite{lin2022vldet}    & \color{gray} CenterNet V2 & \color{gray} Swin-T & \color{gray} 28M  & \color{gray} Box     & \color{gray} 21.9 & \color{gray} 38.6 & \color{gray} 43.6 & \color{gray} 37.7  \\
        VLDet$^\dagger$ \cite{lin2022vldet}        & CenterNet V2 & Swin-T & 28M  & Box     & \underline{24.1} & 39.7 & 43.4 & 38.5  \\
        \midrule
        \color{gray} Base-only                 & \color{gray} Deform. DETR        & \color{gray} Swin-T & \color{gray} 28M   & \color{gray} Box     & \color{gray} 19.1 & \color{gray} 35.9 & \color{gray} 39.9 & \color{gray} 34.5  \\
        \method (Ours)                      & Deform. DETR        & Swin-T & 28M  &  Box    & \textbf{24.3} & 36.9 & 40.8 & 36.3  \\
        \midrule \midrule
        \multicolumn{9}{l}{\textit{Large-scale comparison}} \\
        \midrule
        ViLD \cite{gu2022vild}            & Mask RCNN    & RN152    & 84M & Box+Mask                                   & 19.8 & 27.1 & 34.5 & 28.7  \\
        OWL-ViT \cite{minderer2022owlvit}         & DETR         & ViT-L/14 & 304M & Box                                 & 25.6 & -    & -    & 34.7  \\
        \color{gray} Base-only$^\dagger$ \cite{lin2022vldet}       & \color{gray} CenterNet V2 & \color{gray} Swin-B   & \color{gray}  50M & \color{gray} Box                                  & \color{gray} 27.4 & \color{gray} 44.2 & \color{gray} 48.4 & \color{gray} 43.0  \\
        VLDet$^\dagger$ \cite{lin2022vldet}          & CenterNet V2 & Swin-B   & 50M & Box                                  & \textbf{27.6} & 43.1 & 47.8 & 42.3  \\
        \midrule
        \color{gray} Base-only       & \color{gray} Deform. DETR & \color{gray} Swin-B   & \color{gray} 50M &  Box                                 & \color{gray} 22.0 & \color{gray} 40.2 & \color{gray} 43.5 & \color{gray} 38.4  \\
        \method (Ours)  & Deform. DETR & Swin-B   & 50M & Box                                  & \underline{27.0} & 40.0 & 44.5 & 39.5  \\
        \bottomrule
        \end{tabular}
    }
    \vspace{-0.2cm}
    \caption{
        Open-vocabulary detection results on LVIS datasets. All of APs indicate box AP. \#params denote the number of the backbone parameters.
        * denotes re-evaluation to measure box AP and $\dagger$ denotes re-training with box annotations only.
        The best and second-best results are highlighted using \textbf{bold} and \underline{underlined} numbers, respectively.
    }
    \vspace{-0.1cm}
    \label{table:lvis}
\end{table*}

\section{Experiments}

\subsection{Datasets}
\paragraph{CC3M.}
The CC3M dataset \cite{sharma2018cc3m} contains 3 million image-text pairs collected from the web; each text describes the global context of the associated image.
We use this dataset to train the proposed \method module.

\paragraph{LVIS.} 
The LVIS dataset~\cite{gupta2019lvis} includes detection and instance segmentation annotations for 100K images with 1,203 object classes divided into three categories---frequent, common, rare---based on the number of training images. 
Following ViLD~\cite{gu2022vild}, we repurpose the dataset for the open-vocabulary object detection task; we use the common and frequent classes as base classes $\mathbb{C}_{\text{base}}$ (866 categories) for training and rare classes as novel classes $\mathbb{C}_{\text{novel}}$ (337 categories) for testing.
In evaluation, models predict all categories ({$\mathbb{C}_{\text{base}} \cup \mathbb{C}_{\text{novel}}$) for the test split.
We only use box labels for training and report boxAP, while the dataset provides instance segmentation labels. The primary metric is boxAP for rare categories ($\mathbb{C}_\text{novel}$), denoted $\text{AP}_\text{r}$.

\paragraph{RefCOCOg.}
The LVIS benchmark is suitable for comparing the effectiveness of OVOD algorithms, given that it contains numerous novel object classes for evaluation. However, it may not be appropriate for evaluating the detection capability of arbitrary concepts since LVIS only includes noun concepts. Hence, we provide an additional comparison using the referring expression comprehension (REC) benchmark, localizing text queries that comprise more diverse concepts, such as ``woman with sun glasses'' compared to ``person'' in standard object detection. We compare algorithms using precision over the top-$k$ predictions ($k={1,5,10}$) on the validation set of RefCOCOg~\cite{mao2016regcocog}.

\subsection{Implementation Details}
We use Deformable DETR \cite{zhu2021deformabledetr} for both region proposal extraction and OVOD.
Specifically, we train Deformable DETR as a region proposal network (RPN) on LVIS-base.
Once trained, we extract region proposals from images in LVIS-base using the RPN. 
The objectness scores of these boxes are assigned to their corresponding encoder output. Finally, when generating pseudo-labels, we filter out some proposals if their IoU values with the boxes of base classes are larger than 0.7 or if their objectness score is smaller than 0.2.

Swin-Transformer~\cite{liu2021swin} is used as a backbone of Deformable DETR. Swin-Tiny (Swin-T) is used for all of the ablation studies. We use CLIP of ViT-B/32.
AdamW \cite{loshchilov2018adamw} optimizer is used for every training. For the \method module, the learning rate is set to $10^{-3}$ with a cosine learning rate schedule. We trained \method module for 4 epochs on CC3M dataset with batchsize 8,192 with 1 epoch warmup. The balancing hyperparameter $\lambda_\text{RKD}$ is set to 20. For OVOD models, the learning rate is set to $2\times 10^{-4}$ and decay to $2 \times 10^{-5}$ at 150,000 iterations. The total training iteration is 180,000 with batchsize 32.

\subsection{Open-Vocabulary Detection Results}
The open-vocabulary detection performances in LVIS are compared in \cref{table:lvis}. 
To clearly reveal quantitative performance differences between the methods, we categorize these methods into two groups based on the backbone size: small and large scales. 
We group models with backbone sizes under 30M to small scale and models over 30M to large scale.
For specificity and fairness, we also report re-training results of the previous state-of-the-art (SoTA) method, VLDet \cite{lin2022vldet}, using the official code\footnote{\url{https://github.com/clin1223/VLDet}} with the same backbone and box-level annotations as our model.

\paragraph{Comparison under controlled backbone scales.}
As shown in \cref{table:lvis}, \method shows competitive performance compared to previous state-of-the-art algorithm, VLDet, in rare categories (\apr{}), although the target classes in LVIS predominantly consist of noun concepts, such as ``dog'', ``bench'', or ``necktie''. These results demonstrate the proposed method facilitates accurate learning of noun concepts, not only expanding target concepts.
Since detectors with Swin-T~\cite{liu2021swin} tend to outperform detectors with ResNet-50~\cite{he2016resnet} despite a similar number of parameters, we re-train the previous SoTA method, VLDet \cite{lin2022vldet}, using Swin-T for a fair comparison.
While VLDet and PLAC show similar performance, we emphasize that \method brings much larger performance gain (19.1 $\rightarrow$ 24.3, $+5.2$ \apr) from its baseline (\textit{Base-only}) compared to VLDet (21.9 $\rightarrow$ 24.1, $+2.2$ \apr).

\paragraph{Large-scale comparison.}
In the large-scale experiments, we observe a similar trend where the base-only training of CenterNet V2 \cite{zhou2021centernetv2}, which corresponds to the baseline of VLDet \cite{lin2022vldet}, achieves striking performances.
Due to the strong performance of the baseline, VLDet achieves the best in this large-scale comparison. 
However, it is important to note that most of the performance gap between VLDet and the proposed method is attributed to the baseline performance improvement. When comparing two methods in terms of the performance gain from its \textit{Base-only}, \method ($+5.0$ \apr{}) shows a much larger performance gain than VLDet ($+0.2$ \apr{}). Also, despite using a relatively smaller backbone, our method achieves better performance than other methods, ViLD \cite{gu2022vild} and OWL-ViT \cite{minderer2022owlvit}.

\begin{table}[!t]
    \centering
    \tablestyle{2pt}{1.1}
    \scalebox{1.0}{
    \begin{tabular}{ll|lll}
        \toprule
        
        
        Method  & Backbone & Prec@1 & Prec@5 & Prec@10 \\
        \midrule
        \color{gray} VLDet & \color{gray} RN50   & \color{gray} 31.4  & \color{gray} 44.8  & \color{gray} 55.1 \\
        VLDet & Swin-T & 32.4  & 46.4  & 56.3 \\
        Ours  & Swin-T & 41.2 (+8.8) & 74.1 (+27.7) & 84.3 (+28.0) \\
        \midrule
        VLDet & Swin-B & 35.3 & 50.5 & 61.5 \\
        Ours  & Swin-B & 40.2 (+4.9) & 73.9 (+23.4) & 82.9 (+21.4) \\
        \bottomrule
    \end{tabular}
    }
    \vspace{-0.2cm}
    \caption{
        Zero-shot referring expression comprehension results on RefCOCOg. The proposed method significantly outperforms the previous best method, VLDet.
    }
    \label{table:refcoco}
    \vspace{-0.3cm}
\end{table}
\subsection{Evaluation on Arbitrary Concepts}
While the LVIS dataset is a popular and extensive benchmark for open-vocabulary detection, it is still limited to noun concepts. 
Thus, we compare the OVOD models using the RefCOCOg~\cite{mao2016regcocog} dataset to evaluate the capabilities to address arbitrary concepts. 
As shown in \cref{table:refcoco}, while open-vocabulary detection performances are comparable between VLDet and \method, \method remarkably outperforms VLDet in the referring expression comprehension (REC) benchmark, demonstrating the capabilities of \method to address arbitrary concepts. 
Note that we directly transfer OVOD models to the REC benchmark in a zero-shot manner without any fine-tuning on the REC dataset, to clearly reveal the capabilities of the OVOD models.

\begin{table}[!t]
    \tablestyle{1.8pt}{1.1}
    \subfloat[
        Pseudo-labeling
        \label{table:ablation_pseudo}
    ]{
        \centering
        \small
        {
            \begin{tabular}[b]{@{}l|c@{}}
                \toprule
                Methods & AP$_\text{r}$ \\
                \midrule
                \color{gray} Base-only             & \color{gray} 19.1 \\
                Concept pool          & 15.1 \\
                CLIP image embed  & 22.4 \\
                \method (Ours)             & \textbf{24.3} \\
                \bottomrule
            \end{tabular}
        }
    }
    \subfloat[
        Loss
        \label{table:ablation_rkd}
    ]{
        \centering
        \small
        {
            \newline
            \begin{tabular}[b]{@{}l|cccc@{}}
                \toprule
                 Loss & AP$_\text{r}$ \\
                \midrule
                \color{gray} Base-only             & \color{gray} 19.1 \\
                MSE   & 20.1 \\
                + RKD      & \textbf{24.3} \\
                \bottomrule
            \end{tabular}
        }
    }
    \subfloat[
        Matching
        \label{table:matching_strategy}
    ]{
        \centering
        \small
        
        {
            \begin{tabular}[b]{@{}l|cccc@{}}
                \toprule
                Matching  & AP$_\text{r}$ \\
                \midrule
                \color{gray} Base-only             & \color{gray} 19.1 \\
                Single-stage & 18.6  \\
                Two-stage    & \textbf{24.3} \\
                \bottomrule
            \end{tabular}
        }
    }
    \vspace{-0.2cm}
    \caption{
        Ablation studies for pseudo-labeling methods,  losses, and  matching strategy on the LVIS dataset. Base-only indicates the baseline model trained using base categories without any pseudo-labeling.
    }
    \vspace{-0.2cm}
    \label{table:ablations}
\end{table}
\subsection{Ablation Studies}
\paragraph{Pseudo-labeling methods.} 
\cref{table:ablation_pseudo} examines the impact of different types of pseudo-labeling methods on the LVIS dataset. 
\textit{Base-only} method only employs text embeddings for base categories during training, and the other pseudo-labeling methods extract region proposals using the trained model with the base-only method.
\textit{Concept pool} method measures cosine similarity between the CLIP image embeddings from extracted region proposals and the concept pool used in RegionCLIP~\cite{zhong2022regionclip}, and assigns the most similar concept to the pseudo-label for the given region proposals. In \textit{CLIP image embed} method, the CLIP image embeddings of the region proposals are treated as a pseudo-label. In \cref{table:ablation_pseudo}, the proposed \method pseudo-labeling method significantly outperforms other methods. It is worth noting that the gain of \textit{CLIP image embed} method may be attributed to a weak but still aligned relationship between CLIP image and text embeddings.

\paragraph{Losses for the \method module.}
Table~\ref{table:ablation_rkd} presents the results of the study on the impact of two training losses for \method module: (1) mean-squared-error (MSE), and (2) relational knowledge distillation (RKD).
The results show that using MSE with RKD loss led to outstanding performance on the OVOD benchmark, surpassing the performance of the model trained solely with MSE.
This indicates that RKD loss aids in translating CLIP image embeddings to CLIP text embeddings more accurately.

\paragraph{Two-stage matching.}
\label{sec:ablation_matcher}
Table~\ref{table:matching_strategy} investigates the impact of the two-stage matcher when applying pseudo-labels to train Deformable DETR.
It turns out that the two-stage matcher is critical in terms of boosting OVOD performances.
Experimental results show that even though the loss function for base categories and pseudo embeddings are the same, there must be a priority for applying to the OVOD model.

\begin{figure}[!t]
    \begin{center}
        \includegraphics[width=1.0\linewidth]{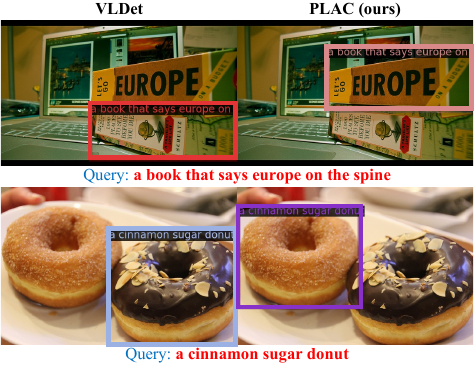}
    \end{center}
    \vspace{-0.3cm}
    \caption{ 
    Visualization of top-1 detected region for VLDet (left) and \method (right) on RefCOCOg dataset. \method is more effective in detecting arbitrary concepts described by free-form text queries.
    }
    \label{fig:sup_fig_text}
\end{figure}

\subsection{Qualitative Analysis}

\paragraph{Alignment between free-form text and region.}
To investigate the capability of localizing arbitrary concepts, we present detection results for the given free-form text using the RefCOCOg~\cite{mao2016regcocog} dataset in \cref{fig:sup_fig_text}.
In the first row in \cref{fig:sup_fig_text}, \method-based OVOD successfully detects the book whose title is matched with the given text query, but VLDet fails.
We conjecture that the pre-defined concept pool employed by VLDet restricts the knowledge contained in the CLIP text embedding, leading to a loss of its potential in the CLIP embedding space, such as for OCR
In addition, PLAC is more effective in understanding fine-level concepts (\eg, cinnamon sugar donut) compared to VLDet that is confused with the chocolate-covered donut, as shown in the second example.

\section{Conclusion}
In this paper, we propose a novel pseudo-labeling method, named \fullmethod (\method), which enables the learning of arbitrary concepts beyond noun concepts in open-vocabulary object detection (OVOD). 
While previous pseudo-labeling methods for OVOD rely on noun parsing to generate pseudo-labels, \method achieves pseudo-labeling beyond noun concepts by learning arbitrary image-to-text mapping.
As demonstrated in our experiments on the LVIS benchmark, the proposed method leads to a significant improvement in OVOD performance, achieving \apr{} of $24.3$ and $27.0$  compared to $19.1$ and $22.0$ achieved by the baseline.
Furthermore, we demonstrate the capabilities of \method in addressing arbitrary concepts by the REC benchmark. In this experiment, the proposed method surpasses the previous best OVOD method, highlighting its potential for advancing research in OVOD beyond noun concepts.


\bibliography{aaai24}

\newpage
\section*{Appendix}

\subsection{Interpretation of PLAC Embeddings}
As described in the method section, we introduce the \method module to learn an arbitrary image-to-text mapping in CLIP latent space.
Specifically, the \method module is learned to map CLIP image embedding ($e^i$) to CLIP text embedding ($e^t$), thus providing a PLAC embedding as a pseudo-label. 
To analyze the meaning of the predicted pseudo-labels (\ie, PLAC embeddings), we train a decoder $D$ that takes a CLIP text embedding $e^t$ and generates a caption $x^t$, \ie, $x^t=\mathcal D(e^t)$;
note that we use the decoder as an interpreter for PLAC embeddings as
PLAC embedding is a predicted CLIP text embedding.
Then, we present examples including 1) input image, 2) ground-truth caption, and 3) generated caption using PLAC embedding in \cref{fig:t2t-examples} from validation set of image-text paired dataset, CC3M~\cite{sharma2018cc3m}.
The examples demonstrate the capabilities of the learned \method module to map arbitrary concepts between images and texts; 
PLAC embeddings (\ie, our pseudo-labels) contain complex semantics of the given image beyond simple nouns, \eg, `\textit{woman}' vs. `\textit{a cartoon style portrait of a woman looking really upset}'.
Consequently, PLAC embeddings facilitate the learning OVOD model by providing the knowledge of CLIP properly.

\subsection{Additional Qualitative Results on LVIS}
We present examples of detection results given by our method in \cref{fig:qual-examples}. Our model effectively detects base categories as well as novel ones in LVIS dataset.

\subsection{Additional Qualitative Results on RefCOCOg}
In \cref{fig:sup_fig_color} and \cref{fig:sup_fig_text2}, we compare the detection results for free-form text queries between VLDet~\cite{lin2022vldet} and the proposed \method-based OVOD. 
Our method predicts better than VLDet for given free-form text queries from two perspectives: (1) recognition of objects' color and (2) alignment between free-form text and region.

\paragraph{Recognition of object color.}
\cref{fig:sup_fig_color} shows the different tendency of predictions between VLDet and the proposed \method-based OVOD.
In the results, we observe that VLDet occasionally fails to distinguish the color of objects described in the given free-form text queries. If the same objects of different colors exist in the image, VLDet's results tend to ignore the objects' color.
For example, in the first-row example in \cref{fig:sup_fig_color}, for the query ``a red color surfboard'', VLDet confuses it with ``white surfboard".
In contrast, the \method-based OVOD shows the ability to correctly distinguish the color of the surfboard. 
These results clearly expose the limitations of the concept pool-based OVOD, which cannot include concepts beyond nouns, \eg, colors. 
For instance, ``blue van" or ``red hydrant" is not included in the concept pool, as they are not common in the dataset.

\begin{figure}[h]
    \begin{center}
        \includegraphics[width=\columnwidth]{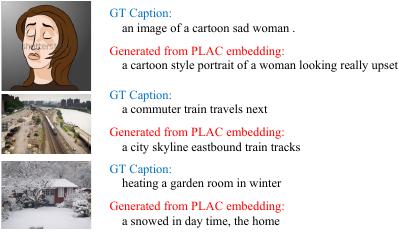}
    \end{center}
    \vspace{-0.2cm}
    \caption{Comparison between ground truth (GT) captions and generated captions from \method embeddings. The generated captions from \method embeddings accurately capture the semantics of images, equivalent to the GT captions.}
    \label{fig:t2t-examples}
\end{figure}

\begin{figure}[h]
    \centering
    \includegraphics[width=\columnwidth]{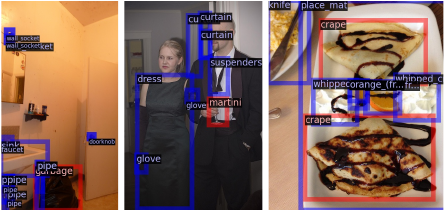}
    \caption{
    Detection results on the validation set of LVIS.
    The base and novel categories are indicated as blue and red boxes, respectively.}
    \label{fig:qual-examples}
\end{figure}

\paragraph{Alignment between free-form text and region.}
The detection results of the different text query with the same given image are presented in \cref{fig:sup_fig_text_com2}.
From an example of two books with distinct titles, the proposed \method-based OVOD demonstrates the capability to identify the book corresponding to a given query. 
Furthermore, when presented with an alternate book title as a query, it exhibits the ability to detect the corresponding new book.
These results illustrate that the proposed \method-based OVOD inherits the OCR capabilities previously exhibited in CLIP.
The other examples also show that the \method-based OVOD responds sensitively to the given queries, and this can serve as evidence that the proposed \method embedding is capable of expressing finer concepts.

\begin{figure*}[t]
    \centering
        \includegraphics[width=1.0\linewidth]{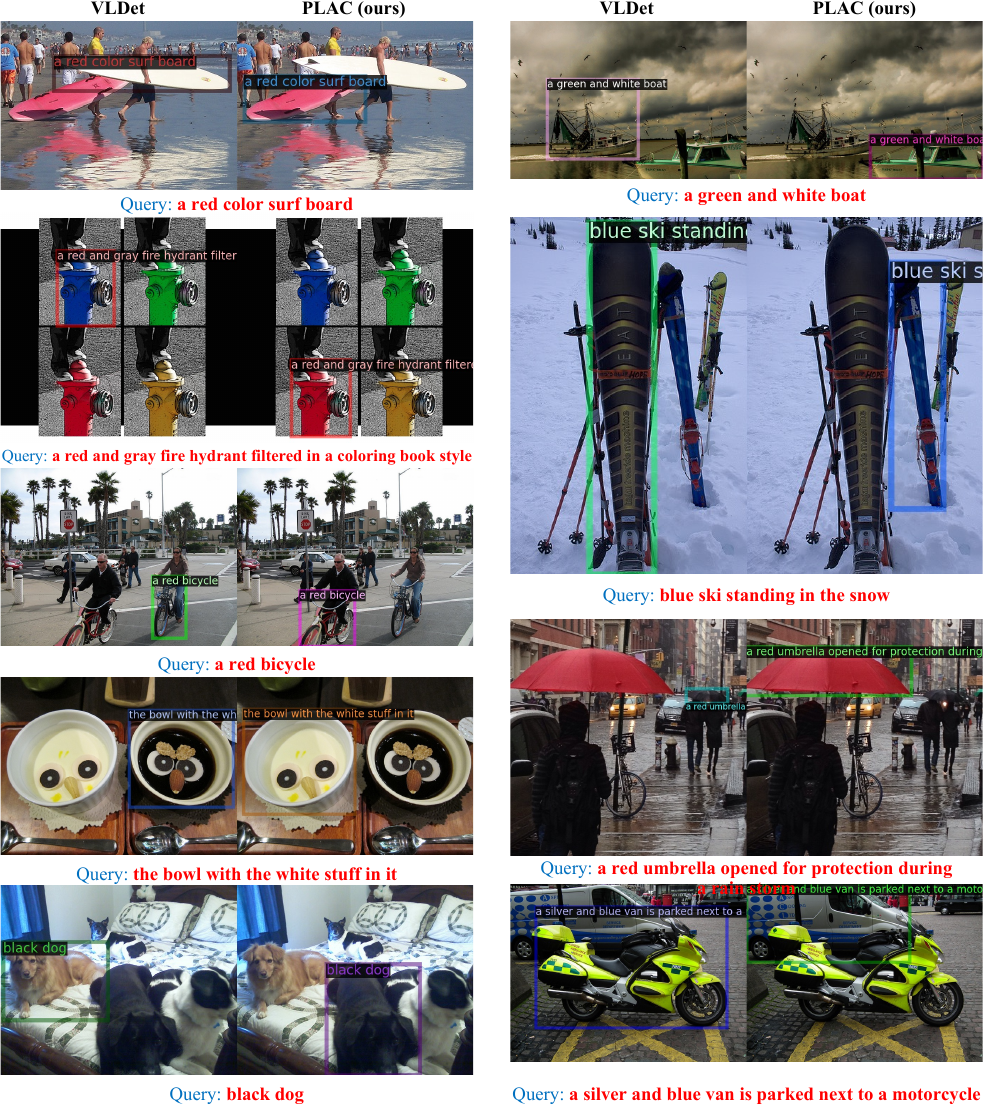}
    \caption{
    \small 
    Visualization of top-1 detected region on RefCOCOg dataset. The left and right of figures indicate the results of VLDet and \method, respectively. The results show that the proposed \method-based OVOD can distinguish the color of objects.
    }
    \label{fig:sup_fig_color}
\end{figure*}
\begin{figure}[h]
    \begin{center}
        \includegraphics[width=1.0\linewidth]{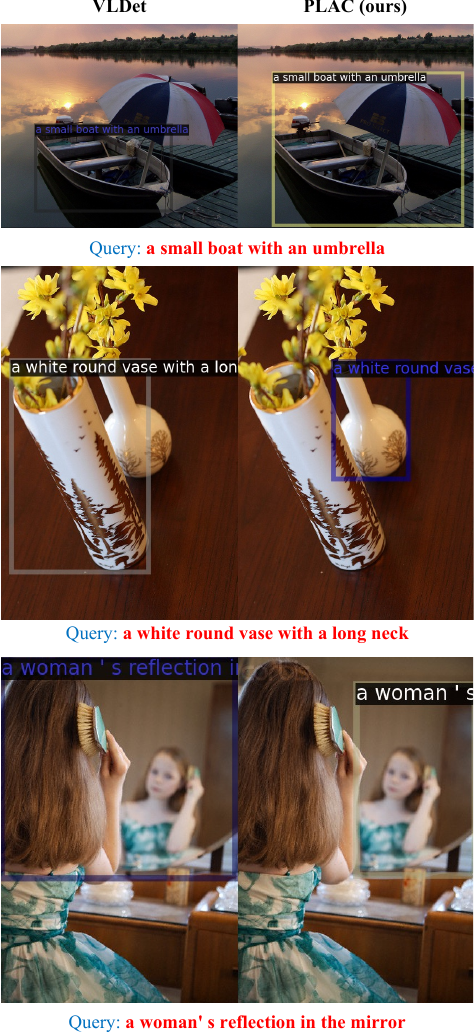}
    \end{center}
    \caption{
    \small 
    Visualization of top-1 detected region on RefCOCOg dataset. The left and right of figures indicate the results of VLDet and \method, respectively. The results show the proposed \method-based OVOD can understand free-form text queries and detect specific objects described.
    }
    \label{fig:sup_fig_text2}
\end{figure}

\begin{figure}[!t]
    \begin{center}
        \includegraphics[width=1.0\linewidth]{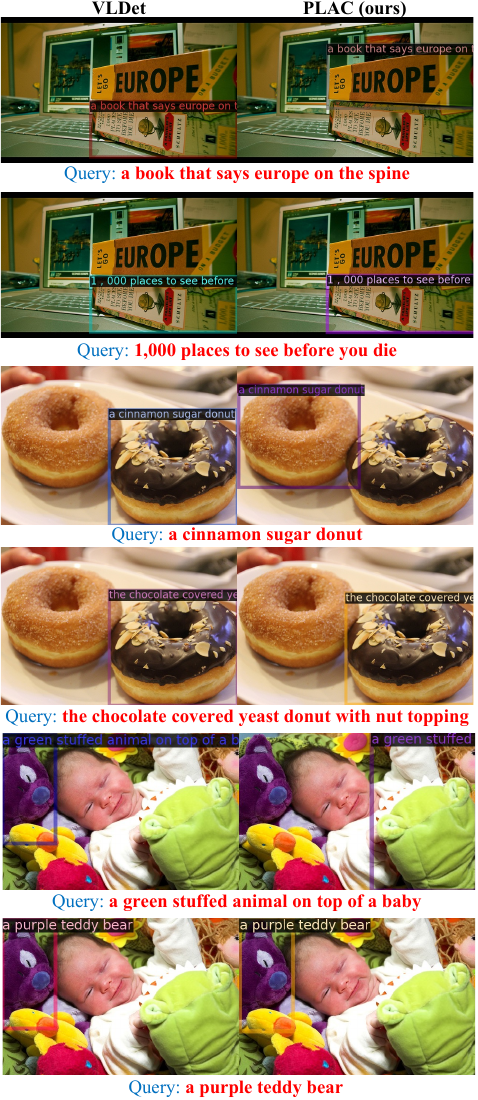}
    \end{center}
    \caption{
    \small 
    The difference in the detection results with the different query text. The \method-based OVOD predicts different regions corresponding to the given text queries.
    }
    \label{fig:sup_fig_text_com2}
\end{figure}

\end{document}